\documentclass[11pt]{article}
\pdfoutput=1
\usepackage[table]{xcolor} 
\usepackage{amssymb,amsmath,amsfonts,eurosym,geometry,ulem,graphicx,caption,color,setspace,sectsty,comment,footmisc,caption,pdflscape,subfigure,array,hyperref}
\usepackage{tikz}
\usepackage{placeins}
\usepackage{multirow}
\usetikzlibrary{shapes.geometric, arrows}

\newcolumntype{L}[1]{>{\raggedright\let\newline\\arraybackslash\hspace{0pt}}m{#1}}
\newcolumntype{C}[1]{>{\centering\let\newline\\arraybackslash\hspace{0pt}}m{#1}}
\newcolumntype{R}[1]{>{\raggedleft\let\newline\\arraybackslash\hspace{0pt}}m{#1}}

\usepackage{tabularx}
\usepackage{enumitem} 
\usepackage{colortbl} 
\usepackage[most]{tcolorbox}
\usepackage{url}
\usepackage{array}
\usepackage{enumitem} 

\definecolor{mygray}{gray}{0.9}

\begin{document}

\begin{titlepage}
\pagenumbering{gobble}
\title{Contextual Confidence and Generative AI\thanks{Corresponding author: shreyjain@microsoft.com.}}
\author{Shrey Jain\thanks{Microsoft Research Special Projects.} \and Zoë Hitzig\thanks{Harvard Society of Fellows.} \footnotemark[4] \and Pamela Mishkin\thanks{OpenAI.}}

\date{October 2023}

\maketitle

\onehalfspacing

\begin{abstract}
Generative AI models perturb the foundations of effective human communication. They present new challenges to \textit{contextual confidence}, disrupting participants’ ability to \textit{identify} the authentic context of communication and their ability to \textit{protect} communication from reuse and recombination outside its intended context. In this paper, we describe strategies – tools, technologies and policies – that aim to stabilize communication in the face of these challenges. The strategies we discuss fall into two broad categories. \textit{Containment} strategies aim to reassert context in environments where it is currently threatened – a reaction to the context-free expectations and norms established by the internet. \textit{Mobilization} strategies, by contrast, view the rise of generative AI as an opportunity to proactively set new and higher expectations around privacy and authenticity in mediated communication.
\end{abstract}
\end{titlepage}

\newpage

\tableofcontents \newpage

\onehalfspacing

\pagenumbering{arabic}

\section{Introduction}

Generative AI is the latest technology to challenge our ability to identify and protect the context of our communications. To see how – and why it matters – it is useful to step back and reflect on the importance of context in communication, and to position generative AI in the sweep of communication technologies that have come before it.   

\vspace{2mm}

Context is what enables effective communication – it is, in a sense, what binds message to meaning. Consider your most recent face-to-face interaction. Perhaps the interaction was with a close friend, family member, or colleague. Perhaps it was with a perfect stranger. Maybe it was a full conversation, or a brief hey-how-are-you, or even just a nod in the hallway. 

Either way, the mere fact that the interaction took place in person likely supplied a common understanding of some basic context. If you or the other person had to summarize the interaction, both of you would likely be able to answer simple questions about where, physically, the interaction took place, on what day, and at what time. Moreover, your answers would likely coincide. You would both be able to describe – at some level of detail – some facts about the identity of the other. Even if it was a perfect stranger, you could likely describe their appearance. If it was someone you're close to, you might have a refined sense of not only who you were speaking with but also their emotional state at that moment. Rich sensory details, combined with our keen faculties for picking up on contextual cues honed over millennia of social, biological and cultural evolution, power our interpretations of face-to-face interactions – from the most rudimentary greeting from a stranger or acquaintance to in-depth conversations with loved ones and colleagues.

Each advance in communication technology flattens or distorts or rearranges, in its own particular way, the contextual markers available in face-to-face communication. The very first writing systems allowed messages to travel across distance and time, and yet divorced the message from its sender and its occasion. So did the printing press, at scale. Then the telegraph, the telephone and the radio. Then emails, text messages and social media. Each of these technologies broadened the possibilities for communication. And at the same time, each is a mediator, and by definition, mediation compresses context. 

In response to each development in communication technology, we develop new norms and expectations to reassert context where it is threatened, preserving our ability to extract intended meaning from the communication. Ancient cuneiform tablets bore a seal impression rolled into wet clay, serving as a signature indicating that a particular individual or institution authored and authorized of the contents of the tablet. Ever since books have been printed, they have included a colophon, detailing facts about the book’s production – the publisher, the date, the city, the press and so forth. Since the early 13th century, paper makers have embedded watermarks in their paper, to signal facts about the quality of the paper or the intended origin and destination of the missive written upon it.\footnote{For an insightful history of communication technologies, see \cite{urquhart2018communication}.} Telegrams, like letters sent through the postal service now, were stamped with information about the originating office, and the date on which they were sent. 

Our efforts to reassert context in communication don't always keep up with the context-compressing technologies, but rather chase a few lengths behind. When Orson Welles, in the early days of radio broadcast, aired a radio adaptation of H.G. Wells' \textit{War of the Worlds} formatted to sound like real-time news bulletins reporting a Martian invasion, some listeners believed the fictional events were real, leading to panic. The extent of the panic was exaggerated by newspapers, including the \textit{New York Times} which reported on a ``wave of mass hysteria" – and attempted to discredit their emergent competitors in news provision: ``The nation as a whole continues to face the danger of incomplete, misunderstood news over a medium which has yet to prove ... that it is competent to perform the news job" \cite{schwartz2015broadcast}. Regardless of the true extent of the hysteria, the episode underscored the importance of clearly establishing context in a medium that carries both fact and fiction. Following the incident, radio hosts were more careful about providing context to their broadcasts, inserting disclaimers frequently – e.g. when the content is fictional, when the content is a re-broadcast, or when the content is paid for by a sponsor.\footnote{The bodies that govern radio broadcasting also developed rules to enforce the use of such disclaimers. For example, the Federal Communications Commission's rules were amended to require that broadcasters disclose when content is sponsored (47 C.F.R. \textsection 73.1212) and to forbid ``the broadcast of hoaxes that are harmful to the public" (47 C.F.R. \textsection 73.1217).} 

We continued our pursuit of context as a vast portion of our communication moved into the digital realm. We invented emojis and deployed them ubiquitously, making our digital communication more expressive through the combination of text and non-text characters. When social media became a leading source of false and misleading information online, we adapted fact-checking measures to this new territory, with tools like Community Notes on X (formerly Twitter), which provide context to otherwise compressed 280 character messages. 

Generative AI is the latest technology to upset context in our communication ecosystem. With its ability to mimic human thought processes and replicate human-like interactions, generative AI arguably introduces an unparalleled layer of contextual ambiguity. Voice scams aren't merely deceptive recordings anymore; they can be highly tailored AI-generated imitations of our loved ones \cite{atleson2023chatbots, verma2023WSJ}. Catfishing, once reliant on static fabricated identities, now has the potential to deploy dynamic AI personas that can evolve in real-time \cite{catfishing}. Misinformation campaigns, previously limited by human cognitive capacities and speed, can be supercharged by AI algorithms churning out vast amounts of manipulative content at lightning speed \cite{kreps2023potential}. The undisclosed usage of AI models in creating content makes it difficult to discern human from machine, jeopardizing the trust and authenticity we once took for granted in education, commerce, democracy and cultural production.

As we trace the trajectory of communication technologies and the standards and norms that have arisen in response, two key facts about generative AI in 2023 come to the fore. First, generative AI, like many communication technologies that came before, confounds our ability to discern context. Second, the norms governing our experience of AI-enabled communications are inchoate. Taken together, these two facts suggest that we currently face a high-stakes opportunity to develop and deploy tools that establish confidence in context. The strategies we adopt (or not) in the near future may determine whether we regress or progress in our ability to communicate – and therefore whether we regress or progress in our ability to lead rich and differentiated social lives.\footnote{One of us has argued elsewhere that generative AI’s threats to communication also constitute threats to the foundations of a “plural” society \cite{pluralpublics}.}  

\subsection{Defining Contextual Confidence}

This paper presents an overview of strategies that deserve our immediate attention if we hope to ensure that generative AI improves – or at least does not frustrate – our ability to communicate. These strategies respond to generative AI’s challenges to what we call \textit{contextual confidence} in communication or information exchange. The context of a communication or information exchange is the “who, why, where, when, and how” of the exchange. Note that context does \textit{not} include “what” the communication is about – the content of the message is distinct from the context, and context informs how participants interpret the content.\footnote{Of course, the separation of content and context we insist on here is an oversimplification. In reality, there may be a rich interplay between the two – a journalist may have a clear sense of context that leads her to trust a particular source's report of some content. But then, if the journalist publishes a statement from the source that turns out to be false, then upon learning that the content turned out to be false, the journalist would in turn have doubts about her understanding of the original context.} 

In settings with high contextual confidence, participants are able to both
\begin{itemize}
\item[]\begin{enumerate}
    \item \textit{identify} the authentic context in which communication takes place, and
    \item \textit{protect} communication from reuse and recombination outside of intended contexts.
\end{enumerate}
\end{itemize}
With greater mutual confidence in their ability to \textit{identify} context, participants are able to communicate more effectively. They can choose how and what to communicate based on their understanding of how their audience will understand the communicative commitments of both speaker and sender \cite{wittgenstein, austin}. With greater confidence in the degree to which their communication is \textit{protected} from reuse or repetition, participants can more appropriately calibrate what, if anything, they want to share in the first place. When all participants in an exchange have contextual confidence,  communication is most efficient and meaningful. This principle mirrors Shannon's source code theorem: messages can be encoded in fewer bits when the receiver will rely on context, as well as the message itself, for decompression \cite{shannon1948mathematical}.\footnote{To illustrate the interplay between the contextual security of a communication medium and the content of our messages, it is instructive to juxtapose work emails with personal texts. A company-provided email system, inherently formal and transparent, could inhibit frankness. Employees often tread carefully, aware that their messages are not truly private and can be surveilled by employers. These emails are also susceptible to being forwarded, quoted out of context, or inadvertently made public. Conversely, personal text messaging appears more safeguarded, fostering a sense of intimacy and discretion. When a colleague opts to send a message via text over a work email, it may not be merely a matter of convenience. This choice also allows the sender to communicate more freely and precisely, and also indicates to the receiver that the content is potentially sensitive. The recipient, recognizing this context, might infer that the shared information should remain confined to that conversation and treated with discretion. Different conditions of contextual security over text message versus work email give rise to different levels of contextual confidence and expressiveness.} 

\subsection{Relationship to Other Frameworks: Contextual Integrity and Information Integrity}

In recent discussions about the social impacts of generative AI, researchers have enumerated areas of concern that include “Privacy and data protections,” “Trustworthiness and autonomy,” “Misinformation,” “Information harms,” “Information integrity,” “Disinformation” and “Deception,” to name a few.\footnote{See, for example, \cite{solaiman2023evaluating, shelby2023sociotechnical, weidinger2021ethical, bommasani2021opportunities, weidinger2022taxonomy, shevlane2023model} and \cite{ brundage2018malicious}.} Contextual confidence collects many such concerns under a single heading, by casting authenticity and privacy as two sides of the same coin. Without an ability to authenticate context, we can’t establish norms for protecting it. Without an ability to protect context, it may be useless to authenticate it. In the face of generative AI,  authenticity and privacy cannot be treated as distinct issues. 

In focusing on the contextual norms and expectations that protect information, we draw heavily on the theory of privacy as “contextual integrity” \cite{nissenbaum2004privacy}. According to the theory of contextual integrity, an information flow is “private” if it conforms with the norms and expectations that govern information flow to a particular, non-universal audience. Privacy is violated when a ``contextual integrity norm" is violated – that is, when information is shared outside its intended context in a way that defies the norms of the original transmission. 

Contextual confidence, like contextual integrity, allows for a discussion of privacy that goes beyond “control over personal information,” and also goes beyond a strict “private” versus “public” binary. Instead, it acknowledges a range of “publics” within which participants should be able to expect and respect some degree of privacy and where that privacy embraces not only control of personal information but control of the range of things that might be the objective of communicative acts.\footnote{In this piece, we discuss strategies that promote contextual confidence in the sense that they help to set norms around the identification and protection of context. When it comes to protecting context, it is important not only that there are norms, but that the norms are \textit{respected}. Indeed, a norm cannot be a norm if there is no expectation that it is respected. However, once a norm is in place, there are many ways for norms to be violated. The strategies we discuss in this paper are primarily concerned with \textit{setting} protection norms, rather than the \textit{enforcement} of protection norms. We discuss the follow-on need for \textit{enforcement} strategies in Section 4.} 

Also like contextual integrity, contextual confidence is both a heuristic framework for determining when needed confidence has been violated and an ideal to which we ought to aspire. We see contextual integrity as an ideal for how communicative commitments to context-specific norms support privacy for \textit{specific acts of communication}; against the standard of that ideal, we can see violations. Meanwhile, contextual confidence communicates an ideal for the \emph{ecosystem of communication}. In an ecosystem with high contextual confidence, there is a low probability of violations of contextual integrity. The aspirational yardsticks of contextual confidence and integrity help us evaluate strategies (policies, tools, and technologies) that promote effective communication. These strategies vary in the way that they remedy challenges raised by generative AI. Some of the strategies we discuss can be directly embedded in generative AI tools, and other strategies we discuss are merely important to invest in as generative AI tools flood our communication ecosystem. 

Our framework of contextual confidence speaks to issues related to authenticity in a way that extends from but builds beyond contextual integrity. For instance, “disinformation,” – false or inaccurate information intended to deceive – may be understood as a violation of contextual integrity; it defies a norm of truthfulness that governs many (but not all) communicative acts. But those norm violations, or rather their frequency, will be a symptom of an information landscape with low contextual confidence, where the guardrails that limit violations are not in place. Disinformation succeeds when people struggle to identify the genuine context from which a statement emerges – political propaganda, for example, obscures features of context like \textit{who} is really behind some message (a political interest), and \textit{why} the message has reached a particular recipient (because the recipient is a susceptible target for persuasion).

Note that contextual integrity and contextual confidence both deal with disinformation in a way that is distinct from other frameworks like \textit{information integrity} \cite{informationintegrity}. Information integrity refers to the reliability and accuracy of information. It suggests that there is a universal benchmark of “accuracy,” and that the public ought to trust high quality information which meets this benchmark.  

But the norm that stabilizes communication is not truth but \textit{truthfulness} – a commitment to communicating only the truth as best as one understands it \cite{allen2018democratic}. A breaking news story may contain statements that are presented truthfully one day – that is, those facts are believed to be true by those who publish them – but then they may be shown in fact to be false the next day. Or truthful but partial accounts may be offered in one context, while people operating in other contexts, when presented with the same facts, might consider them untruthful on the grounds of omission.  For instance, in the realm of social media, individuals may want to share certain facts about themselves to one group of people and a wholly different set of facts to another. That social media makes it difficult to authentically present oneself differently to different audiences is a symptom of what internet and media scholars call “context collapse” \cite{marwick2011tweet, baym2012socially, baym2015personal}. Norms of truthfulness can operate and sustain the public sphere, alongside discrepancies and conflicts in information. Supporting contextual confidence requires ascertaining how to help people navigate informational inconsistencies; this navigation is more straightforward when norms of truthfulness, and other allied concepts, can be relied upon.

\subsection{How Generative AI Fits in the Evolution of Communication Technologies That Challenge Context}

As discussed above, it is much easier to avoid context collapse – and to ensure high contextual confidence – in face-to-face communication. In in-person communication, parties in the interaction are able to more easily \textit{identify} context. They can read some basic facts about: who they are talking to (even if they are talking to a stranger), why they entered into the exchange, where and when the communication is taking place. They also understand how they are communicating – in a particular language, body language, dialect, tone, affect and manner. With this context, parties can quickly establish the most efficient way to communicate with one another. For example, consider two strangers who meet and begin to speak to each other in English – one notices that the other speaks with an accent from their home country, and so they switch to their native language, enabling a more expressive conversation. The act of identifying context is the development of a hypothesis about the what, who, when, and why – in this case, the hypothesis includes the sharing of a native language. The act of validation or authentication involves confirming the accuracy of that hypothesis. In this example, the hypothesis is confirmed – and the identification converted into authentication – when the interlocutor responds fluently in the hypothesized native tongue.

In face-to-face communications, features of the context often determine strong norms or expectations about how the communication is \textit{protected} – i.e. what can and can’t be shared outside of the intended context. For example, close friends who meet periodically to work through details about their personal lives probably assume that what they share will not be repeated widely to others in their network. If one friend recorded the other without their consent, the recorded friend would probably feel that a friendship norm had been violated (and in many places, including twelve states in the U.S., such a recording would be illegal). When being interviewed by a journalist, it is assumed that all statements may be reproduced as quotes in an article unless explicitly classified as “off the record” or “on background.” When there is no obvious norm, for instance at the first meeting of a newly formed interest group, it is common to establish a protective norm. The group may clarify that everything said in the room stays in the room, or the group may choose to follow a looser variant of such a protective norm like the Chatham House Rule. 

When communicating from a distance, by contrast, it is harder to establish and protect contextual confidence. Conversing through text, audio or even video alters the rich sense data available that identifies context in a face-to-face interaction. Strangers who begin conversing through text in English may take time to discover that they share the same native language  – whereas in person, this can be detected immediately.  Online and phone scams target our struggles to identify context. An email that appears to be from a colleague could be from a hacker who compromised their account. A hurried phone call from a family member in trouble could be an AI-generated imitation of the family member, plus a spoofed caller ID.

In addition to the challenges in identifying context, there are a few features of mediated communication that present extra challenges to the protection of context. First, there is often an intermediary – many of our communications travel through intermediaries like messaging services, telecom providers, and social media platforms. To what degree are communications protected from reuse or recombination by the intermediary? Second, the communications are already packaged up for travel, so they can be instantaneously shared outside of their intended context, and often more verifiably so than in-person communications. In instants, an email can be forwarded, a series of text messages can be screenshotted, and a photo, video or audio recording can be sent onward to an unintended recipient. These forwarded artifacts purport to be firsthand facsimiles of the original communication. Compare this lack of protection against digital reuse to the natural protection of in-person conversations – the basic fact that secondhand reports are often met with a presumption of inexactness helps to protect context in face-to-face conversation.\footnote{Suppose Shrey says something to Zoë in an in-person meeting. Zoë may repeat what Shrey said to Pamela, but Pamela will assume that Zoë is not repeating Shrey's statement verbatim – and she may wonder whether Zoë has faithfully summarized or otherwise edited the statement in passing it on. Compare this to the equivalent digital communication: if Shrey writes something to Zoë in an email, Sebastian may forward the email to Pamela. In this case, Pamela assumes that Shrey wrote exactly the text shown in the forwarded email.} 

Some of the defining policies, laws and technologies of the last decade can be understood as attempts to reinstate context in mediated communication. Privacy laws like the General Data Protection Regulation (GDPR) in Europe and similar legislation in California aim to safeguard user data and consent. There have been noteworthy pushes for transparency, like the EU AI act, which, for example, demands clarity on why users receive particular content recommendations in recommender systems. In tandem, we've seen the rise of authentication tools designed to validate the authenticity of communication sources; examples include DKIM and DMARC for emails, verified badges and content moderation mechanisms on social media platforms. There has been a surge in the adoption of messaging services that provide end-to-end encryption, such as Signal, WhatsApp, and Telegram. Despite these strides, the path ahead towards contextual confidence in mediated communications remains long and complex.

And now, generative AI has thrown a spate of new obstacles onto the path that grow out of and extend beyond the digital communication challenges that have defined the last few decades. Generative AI causes a new set of problems in \textit{identifying} “who” is engaged in communication: one might be communicating with a model or a human.\footnote{The overwhelming focus on making generative AI ``human-like" \cite{brynjolfsson2022turing} suggests that AI's impersonation capabilities will continue to evolve rapidly.} As models and specialized versions of those models proliferate, there will be a further challenge of distinguishing among different models. For example, a doctor may believe they are communicating with a specialized model trained for clinical use but in fact they are engaging with a model that a malicious actor installed in its place. Furthermore, generative AI models can convincingly imitate specific people or specific types of people. These imitations further challenge our ability to identify “who” we are exchanging information with, as well as our ability to accurately make inferences based on “how” (style, tone, manner) information is conveyed to us.

To see how generative AI also strains our ability to protect context, one need look no further than the training process itself. At the most basic level, many of us who shared text, images, likenesses, and other content on the internet in the last two decades did not expect, at the time we shared it, that these data shared in one context (the internet) would be used in a wholly different context (to train a generative AI model). Most of the data used to train the prominent generative AI models was reused, in some sense, outside of its intended context – going forward, it will be important to highlight which communications are used as part of training data for what (general or specialized) model.  Not only was our data reused outside its intended context, but now an entity may be communicating about us – the AI model – an entity with which we never formally signed up to be in a communicative relationship. A similar thing might be said about cookies. Cookies communicate to advertising programs about us, and then we receive ads. New options to manage our cookies actually give us the chance to determine what communicative relationships we wish to be part of in the first place. Ideally, we would choose those communicative relationships. And even with those choices there are further questions: What could be inferred about us from the synthesis of various facts about us and our networks? How does the model represent or misrepresent me or the groups and cultures to which I belong?

\subsection{Overview of the Paper}

As a framework, contextual confidence inextricably links privacy and authenticity, and clarifies the impact of new technologies on communication. By precisely locating challenges to the identification and protection of context, we can better understand which mitigation strategies are worthy of pursuing and prioritizing. While this paper focuses on generative AI, we note throughout that some of the challenges launched by generative AI are simply turbo-charged iterations of challenges that have existed for decades and in some cases millennia. Other challenges we discuss are different not just in degree but in kind. These challenges arise in interactions between generative AI systems and their users and their audience, and yet they have repercussions for the broader communication landscape. Indeed, a feature of contextual confidence is that it is contagious – low contextual confidence in interactions that involve AI systems generate low contextual confidence more broadly, even in forms of communication that do not involve generative AI.

\section{\Large {Challenges to Contextual Confidence from Generative AI}}\label{sec:challenges}

We begin by outlining a set of challenges that will arise in a communication landscape where generative AI models are widely used and available. Many of these challenges are continuations of challenges that already existed with the rise of the internet, while others are more specific to generative AI. The first five challenges we discuss are challenges, broadly understood, to the \textit{identification} of context (Section \ref{sec:challenges_identify}), while the second six challenges, broadly understood, have to do with the \textit{protection} of context (Section \ref{sec:challenges_protect}). We close this section with a brief discussion of how these challenges are heterogeneous in their impacts on different segments of the population (Section \ref{sec:heterogeneity}). The first row of \autoref{table:challenges_strategy} summarizes the challenges discussed.

\renewcommand{\arraystretch}{1.5}

\begin{table*}[h]
    \centering
    \footnotesize 
    \begin{tabular}{|p{4cm}|p{5.9
    cm}|p{5.9cm}|}
        \hline
        & & \\
        &\multicolumn{1}{c|}{ \textbf{ \large  Identifying Context}} & \multicolumn{1}{c|}{\textbf{ \large Protecting Context}} \\
        & & \\
        \hline
        
       \multirow{8}{4cm}{\\ \centering \large\textbf{Challenges}}  & 
            \begin{itemize}
                \item Impersonate the identity of an individual 
                \item Imitate members of social or cultural groups
                \item Create mimetic models of oneself
                \item Imitate another AI model                
                \item Falsely represent grassroots support or consensus (astroturfing)
            \end{itemize}
         & 
        \begin{itemize}
            \item Disseminate content in a scalable, automated, and targeted way
            \item Fail to accurately represent origin of content
            \item Misrepresent members of specific social or cultural groups
            \item Leak or infer information about specific individuals or groups 
            \item Pollute the data commons
            \item Train context-specific models without context specific restrictions
        \end{itemize} \\
        \hline \hline 
        \multirow{4}{4cm}{\large\\ \centering \textbf{Containment}\\ \vspace{2mm}\textbf{Strategies}
        } &
        \begin{itemize}
            \item Content provenance
            \item Community Notes
            \item Centralized digital identities
            \item Identity as a social intersection
        \end{itemize} &
        \begin{itemize}
            \item Usage and content policies
            \item Rate-limiting communication
            \item Prompt protection \& interface design
            \item Data verification
        \end{itemize} \\
        \hline
        \multirow{4}{4cm}{\large\\ \centering \textbf{Mobilization}\\ \vspace{2mm}\textbf{Strategies}} &
        \begin{itemize}
            \item Watermarking
            \item Model verification
            \item Relational passwords
            \item Collusion-resistant digital identities
        \end{itemize} &
        \begin{itemize}
            \item Contextual training
            \item Deniable messages
            \item Data cooperatives
            \item Secure data sharing mechanisms
        \end{itemize} \\
        \hline
        
    \end{tabular}
      \begin{minipage}{0.6\linewidth}
          \vspace{3mm}
    \caption{Challenges to contextual confidence from generative AI (\autoref{sec:challenges}); containment and mobilization strategies for responding to challenges (\autoref{sec:strategies}).}
     \label{table:challenges_strategy}
     \end{minipage}
\end{table*}

\FloatBarrier

\subsection{Challenges to Identifying Context}\label{sec:challenges_identify}
We begin by enumerating five challenges that are primarily directed at the identification of context. Many of these challenges have to do with various kinds of impersonation -- of another person, of a member of a specific group, of oneself (Challenges I\#1-3) -- but we treat them as distinct because each instance raises distinct concerns and opportunities for remedy. We also discuss imitation of other AI models (Challenge I\#4), and imitations of broad publics, as in astroturfing (Challenge I\#5). 
\begin{tcolorbox}[colback=gray!5!white,colframe=black!75!black]\footnotesize
\textbf{Challenge I\#1:} Impersonate the identity of an individual without their consent and in a scalable, automated, and targeted way.
\end{tcolorbox}

\noindent  Generative AI is now able to impersonate specific individuals with high fidelity, in text, audio, and even video. Malicious actors can impersonate specific individuals without their consent, in order to deceive others. Doing so undermines contextual confidence by impairing the target’s ability to identify context, particularly the “who” that they are communicating with. Attackers might leverage high-fidelity impersonations in social engineering attacks, deceiving victims to extract financial resources \cite{atleson2023chatbots, verma2023WSJ, milmo2023paedophiles, horvitz2022horizon}. They may also use such impersonations to make identity theft and identity fraud cheaper and more convincing.

\begin{tcolorbox}[colback=gray!5!white,colframe=black!75!black]\footnotesize
\textbf{Challenge I\#2:} Imitate members of specific social or cultural groups.
\end{tcolorbox}

\noindent Just as generative AI models can convincingly imitate specific individuals, they can also convincingly imitate members of specific social or cultural groups. This possibility for imitation presents opportunities for overcoming various forms of entrenched discrimination alongside new challenges to the identification of context. We focus here on the challenges, while acknowledging that this issue's complexity extends far beyond the scope of this paper.\footnote{Consider, for example, the AI startup erasing call center worker's accents so that they sound like white Americans \cite{callcenter}. Is this practice aiding cultural biases, or fighting them?} For example, when indigenous activists discovered that Whisper (OpenAI’s speech recognition tool) could translate Maori, having been trained on thousands of hours of Maori language, they voiced concerns \cite{chandran2023indigenous}.\footnote{Here, we discuss concerns related to the identification of context raised by this incident, but other concerns centered on data abuse and cultural appropriation, issues which are broadly about the protection of context (see Challenge P\#3 below).} Before language models could imitate a specific social or cultural group with unprecedented fidelity, it was much harder for outsiders to present themselves as members as a specific social group. So, it was easier for members of the group to be stewards of their own culture, with some ability to steer internal and external perceptions of their group toward an authentic understanding. By contrast, outsiders who present themselves as members of the group using the language model may not be attuned to biases in the model, and may thus skew internal and external perceptions, reinforcing stereotypes and distorting the imitated culture. 

\begin{tcolorbox}[colback=gray!5!white,colframe=black!75!black]\footnotesize
\textbf{Challenge I\#3:} Create mimetic models.  
\end{tcolorbox}

\noindent A mimetic AI model is a model that is trained to act as a particular individual in specified scenarios. In the near future, many may delegate their communications via email and text to mimetic models (autocomplete, autocorrect, and other “smart compose” technologies serve as early versions of this). It will soon be possible to send a mimetic model to a meeting to participate on an individual's behalf. As it stands, there are few norms governing their use and disclosure of their use, and yet mimetic models will soon disrupt social norms and expectations. Mimetic models can erode our ability to identify context,  by obscuring “who” is communicating, and “how.” In addition, they may strain our ability to protect context – as information received by a mimetic model may be used in training or feedback by the model itself or by an organization that trained the model in the first place.\footnote{For an overview of ethical issues related to the use of mimetic models, see \cite{mcilroy2022mimetic}.} 

\begin{tcolorbox}[colback=gray!5!white,colframe=black!75!black]\footnotesize
\textbf{Challenge I\#4:} Imitate another AI model.    
\end{tcolorbox}

\noindent As specific industries and organizations develop specialized AI models for particular use cases, it will be possible for malicious actors to imitate one AI model with another via social engineering strategies. This possibility may be particularly threatening in high-stakes industries like healthcare. For example, a malicious actor connected with a pharmaceutical company could imitate or infiltrate a clinician-approved model to push a specific drug. Such possibilities present new challenges to the identification of context – how can users be sure that they are interacting with the intended model, and not a fraudulent one installed in its place?\footnote{Of course there are natural limits to imitation -- as generative AI models are not deterministic, an imitative model's outputs will never perfectly coincide with the target model.}

\begin{tcolorbox}[colback=gray!5!white,colframe=black!75!black]\footnotesize
\textbf{Challenge I\#5:} Falsely represent grassroots support or consensus (astroturfing).
\end{tcolorbox}

\noindent Generative AI makes it cheaper to generate and disseminate messages in support of or in opposition to  particular political opinions or consumer products and services. Generative AI can help to make these messages appear as if they come from a wide range of individuals, creating the appearance of grassroots consensus. A recent study showed that state legislators couldn't tell the difference between AI-generated and human-written emails, illustrating how generative AI may destabilize political communication \cite{kreps2023potential}. Astroturfing obscures the “who” and “why” of communication – making it appear as though many disparate individuals came to a platform to express their genuine opinion, when in fact, there may be a single actor behind the messages, who is disseminating the messages with the specific intention of convincing people to take on a particular position or to take a particular action. Astroturfing, especially when powered by generative AI, can harm consumers, amplify social and political polarization, suppress dissent, and strengthen authoritarian regimes.\footnote{\cite{goldstein2023generative} provides a detailed survey of how generative AI can power influence operations.} 

\subsection{Challenges to Protecting Context}\label{sec:challenges_protect}
Next we turn to challenges that primarily have to do with protecting context. Some of these challenges are continuations of challenges brought on by the internet and predictive AI, and are now exacerbated by the rise of generative AI: for instance, disseminating content in a scalable, automated and targeted way (Challenge P\#1) and failing to represent the origin of content (Challenge P\#2) fall into this category. Other challenges in this category have to do with the generative aspect of generative AI, like Challenges P\#3-P\#6.

\begin{tcolorbox}[colback=gray!5!white,colframe=black!75!black]\footnotesize
\textbf{Challenge P\#1:} Disseminate content in a scalable, automated, and targeted way.
\end{tcolorbox}

\noindent Generative AI can produce and disseminate content on a large scale, automating processes that would otherwise require extensive human effort. It can create hundreds of variations of a given article, each tailored to a different demographic or psychographic profile, and send these out in a fraction of the time it would take human writers and even the prior generation of (non-generative) AI tools like predictive analytics. Any particular statement, image, or video can be lifted out of its original context and incorporated into new content that may be specifically designed to elicit a particular kind of response from its target audience. This capability of generative AI can turbocharge the generation and spread of disinformation, misinformation, propaganda and fraud. Despite significant efforts to align, red-team, and monitor generative AI models, challenges with hallucinations and sycophancy are still apparent as of 2023 \cite{sharma2023towards}, which is an increasingly significant concern as humans are becoming over-reliant on these systems \cite{vasconcelos2023explanations}.

\begin{tcolorbox}[colback=gray!5!white,colframe=black!75!black]\footnotesize
\textbf{Challenge P\#2:} Fail to accurately represent the origin of content.
\end{tcolorbox}

\noindent A defining feature of generative AI is that it is a black box: it recombines vast amounts of data from a wide range of sources, without tracing exactly how it is drawing on different sources in its provision of new content. It has thus been built in a way that makes it difficult to trace where content originated – the very point of generative AI is to synthesize data across many original contexts for reuse in a new context.\footnote{There is a large and promising line of research on data attribution, which strives to track model outputs back to training data. This line of research is further discussed later in the paper, in the section on data cooperatives (Section \ref{sec:datacoops}). } When participants cannot be sure that their communications and other content will be attributed to them, they may be wary of communicating or producing content in the first place. Indeed, the origin-less nature of generative AI has initiated heated conversations about intellectual property and fair use.\footnote{For an overview of the current conversations around generative AI and fair use, see \cite{henderson2023foundation}.}

\begin{tcolorbox}[colback=gray!5!white,colframe=black!75!black]\footnotesize
\textbf{Challenge P\#3:} Misrepresent members of specific social or cultural groups. 
\end{tcolorbox}

\noindent Generative AI models are trained on data that includes cultural expressions from a variety of groups. They are also trained on datasets that include stereotypes of certain groups, as well as instances of cultural appropriation. These groups are therefore limited in their ability to protect the context of their communications -- there may be aspects of their language or culture that they would not have revealed if they knew that these aspects would be used in a wholly different context, i.e. to train a model. When the model recombines these elements into new content, it may replicate harmful stereotypes, engage in cultural appropriation, or fail to respect specific traditions. If the model is used in a commercial setting, it may also enable an outsider to profit from the intellectual or artistic heritage of another group without their consent. These issues closely relate to the issues discussed in P\#2 and I\#2 and yet are sufficiently distinct to warrant their own discussion. Consider again the discussions about Whisper’s Maori translations. Indigenous activists have not only raised concerns about how Whisper may lead to cultural distortion (discussed in I\#2), but have also raised concerns about cultural appropriation, and the extent to which outsiders might misrepresent or misuse cultural narratives through generative AI.

\begin{tcolorbox}[colback=gray!5!white,colframe=black!75!black]\footnotesize
\textbf{Challenge P\#4:} Leak or infer information about specific individuals or groups without their consent.
\end{tcolorbox}

\noindent As generative AI models engage users and synthesize data in unprecedented ways, they may leak or infer data without the consent of data subjects. This may occur because AI models are trained on datasets that include sensitive information and produce outputs that inadvertently reveal these details. Or it may be the case that generative models, with their ability to systematically analyze disparate sources of data, create an ability to re-identify individuals or groups who contributed to training data, or to infer private information through educated guesses. These possibilities for leakage or inference of sensitive information magnify the threats of identity theft and fraud, and also may enable sophisticated social engineering or blackmail schemes \cite{atleson2023chatbots, verma2023WSJ}. Additionally, generative AI models are susceptible to adversarial attacks leading to leakage of of personally identifiable information (PII) \cite{carlini2021extracting, nasr2023scalable}.

\begin{tcolorbox}[colback=gray!5!white,colframe=black!75!black]\footnotesize
\textbf{Challenge P\#5:} Pollute the data commons.
\end{tcolorbox}

\noindent As generative AI models continue to blend and repurpose data outside their explicitly intended contexts, they feed the “data commons” with their output. To the extent that future generations of AI models are trained on these updated “data commons,” they will be even more divorced from context than the original models. This dynamic may lead to ``model collapse," a situation where AI inaccurately represents or loses the ability to represent the complexity or nuances of specific contexts \cite{shumailov2023curse}. As more people come to rely on generative AI models to communicate, the threat of model collapse becomes a threat to effective communication. 

\begin{tcolorbox}[colback=gray!5!white,colframe=black!75!black]\footnotesize
\textbf{Challenge P\#6:} Train a context-specific model without context-specific restrictions.
\end{tcolorbox}

\noindent Organizations, individuals and industries are training context-specific models. If these context-specific models are not accompanied by corresponding restrictions, there will be new challenges to the protection of context. 

There are at least two issues that come about when a context-specific model is trained without context-specific restrictions. First, information used to train the context-specific model may be reused or recombined in a context other than the one intended, enabling third-party access to sensitive information. Thus, this challenge may lead to instances of Challenge P\#4.

Second, some models used for narrow applications – such as a diagnostic tool for clinicians – may produce outputs that can only be correctly interpreted by a narrow class of users. If a user outside this narrow class accesses the outputs of the diagnostic model, they may be badly misled in a way that could produce harm to themselves or others. In particular, if a patient without medical training gains unrestricted access to a diagnostic model, they may self-diagnose incorrectly, rather than visiting a doctor, putting their health at risk.\footnote{Related issues are discussed in the context of Google's Med-PaLM 2 in \cite{singhal2023towards}.}

\subsection{Heterogeneity in Impacts of Generative AI on Contextual Confidence}\label{sec:heterogeneity}

Even in face-to-face communications, different populations vary in their capacity to identify and protect the context of their communications. For instance, people speaking a second language may not pick up on the same nuances and details as someone speaking in their first language. Those with disabilities have different forms of access to -- and understanding of -- contextual information than those without. 

Generative AI can help to overcome some of these obstacles to contextual confidence in face-to-face and digital communication. For instance, for non-native speakers, the technology can generate translations or paraphrases that consider cultural nuances, going beyond literal translation. These tools can ensure that the intended meaning is conveyed without the loss of essential details. For those with hearing impairments, generative AI can generate real-time captions or transcriptions that not only convert speech to text but also identify and highlight key contextual elements, such as the speaker's tone or the mood of a conversation. For visually impaired individuals, generative AI can produce descriptive texts of images, photos, or video content, capturing more than just objects but also their interrelationships, the emotions they might evoke, and other subtle context markers. By catering to these specific needs, generative AI ensures that each individual can engage with information in a way that is both meaningful and contextually rich.

At the same time, generative AI may exacerbate other varied challenges to contextual confidence, and create new ones for different populations.\footnote{See \cite{eudisability} for a discussion of recommendations on how generative AI can be developed via an inclusive ``Design for All" approach.} Populations with limited digital literacy or who might otherwise struggle to navigate shifting technological landscapes will be more vulnerable to the challenges we've discussed. Scams and fraudulent schemes disproportionately target elders \cite{ic3elderfraud2022}, and there are already examples of attackers using generative AI for these schemes \cite{puig2023scammers}. Social norms that emphasize trust over skepticism can further increase susceptibility, as can economic conditions where promises of quick financial gains are particularly tempting. Veterans and military retirees are at higher risk of targeted scams, often through purported benefits or entitlements \cite{cfpb}. Those who are already disproportionate targets of harassment online -- including women, adolescents and the LGBTQ+ community -- may continue to be disproportionately impacted by the automated production of harmful content  \cite{informationintegrity}. 

Although we will not enumerate in further detail how different populations are more or less affected by the challenges we've described, we acknowledge the importance of thinking through the heterogeneous risks so that contextual confidence strategies can be properly matched to the groups under threat. Some groups are particularly vulnerable in the present and immediate future – such as older populations – and considering the usability of contextual confidence strategies for specific populations is critical.

\section{Strategies to Promote Contextual Confidence}\label{sec:strategies}

One possible response to the threats outlined in \autoref{sec:challenges} is to retreat from digital communication. Indeed, a shift back toward in-person communications can mitigate many of generative AI's threats to contextual confidence. But rather than retreat in response to these challenges, one alternative approach is to view the challenges we've outlined as an occasion to develop new tools and policies to promote contextual confidence in \textit{all} forms of communication. As such, the remainder of this section centers on more ambitious strategies for mitigating the challenges outlined in \autoref{sec:challenges}. The strategies that we discuss here fall into two categories. 

First, there are \textit{containment} strategies.  These are strategies that aim to protect and identify context in settings where it is currently threatened. Containment strategies are, to some extent, reactive – they aim to reassert the importance of context in digital communications, countering the expectations that the internet gave us. Containment technologies help to override norms and expectations that were established somewhat arbitrarily in the early days of the internet.\footnote{Many have written about the competing visions that dictated the development of the internet, and how it led to the norms and expectations we have today, see for example \cite{plurality2023association}.} For instance, tools for tracking the provenance of content are containment strategies because they aim to override a norm governing information exchange that has been set in place – the norm that content on the internet need not be traceable to its origin.  The containment strategies we discuss below are largely strategies that already exist or are in development.

The second set of strategies fall under the heading of \textit{mobilization}. Where containment strategies are reactive in the face of challenges to contextual confidence, mobilization strategies are proactive. They work to establish context in novel settings that have arisen or will arise with generative AI. In doing so, these strategies respond to an opportunity to establish new norms and expectations of information exchange. For an example of a mobilization technology, consider watermarking – a tool that flags content (image, text or video) as having been generated by AI. This tool mobilizes confidence in that it offers an opportunity to set a better norm in a new form of information exchange, embedding context in generative AI-enabled communications. In contrast to containment strategies, many of which are already in discussion and development as responses to issues that existed before generative AI, many of the mobilization strategies we discuss here have not been developed or deployed at a meaningful scale.

To recap, containment strategies are reactive while mobilization strategies are proactive. To some extent, this implies that containment strategies and mobilization strategies differ in their novelty, with mobilization strategies being more novel and less developed. But the key distinction between them is not their novelty but whether they respond primarily to existing challenges or to challenges that have not yet become widespread.

Some of the strategies we discuss can be integrated with generative AI products and development (like content provenance and contextual training), and other strategies discussed here are simply important to have in place in a communication landscape in which generative AI tools are widely available. In addition, the strategies we discuss can serve different (non-exclusive) roles. First, these strategies can require the identification and protection of context in isolated settings, deterring behaviors that fail to identify and protect context by creating a nuisance. For example, deniable messaging does not make it impossible to share messages outside of context -- rather than taking a screenshot of a given message, a receiver could take a photo of the message with another phone -- it merely makes it more difficult to share a message out of context. But, second, these strategies can also serve a broader role, in that they help to shift communicative norms toward norms that encourage more protection and identification of context by default. From this perspective, deniable messaging can help shift toward a world where people are less likely to take a photo with another phone out because there is a strong norm around the protection of context -- a photo of a message taken from another phone would be viewed  as a norm violation, and perhaps viewed with suspicion.

In the remainder of this section, we discuss a range of specific strategies that help to contain and pre-empt AI’s challenges to contextual confidence. The second and third rows of \autoref{table:challenges_strategy} enumerate the specific containment and mobilization strategies we discuss. At the end of each subsection discussing a specific strategy, we summarize the challenges from \autoref{sec:challenges} that are addressed by the strategy in question.

Our discussion is not intended to be exhaustive. In fact, our hope is that by offering a framework for understanding the strategies that were apparent to us, we make it easier for others to build out and situate other strategies that we have not discussed. In addition, the value of these strategies requires further evaluation in unique combinations, as their blind use may even undermine their intended function \cite{xiao2023account,akhawe2013alice}.

\subsection{Containment Strategies to Identify Context}

\subsubsection{Content provenance}

Content provenance tools trace information from its origins through its lifecycle. The leading standard for content provenance is the Coalition for Content Provenance and Authenticity (C2PA) \cite{c2pa2023overview} standard. This standard grew out of Project Origin \cite{originproject2023} and the Content Authenticity Initiative \cite{contentauthenticity2023} provenance projects. The origins of content and a history of its alterations are tracked through a ``manifest," a file that is digitally signed by the cryptographic keys of each entity that modifies the content. Content provenance solutions rely heavily on digital identity attestations (as discussed in section \ref{sec:centralized_id} and section  \ref{sec:social_id}) to ``tag" the originator of content. But content provenance goes further, providing not just a ``tag" on content indicating from whom the information originated, but also verifiably tracing content along with all of its alterations as it moves from party to party.  

Existing implementations of the C2PA standard as seen with Project Origin require media companies to trace the origin of content from media capture on a hardware device through to publication online. Dependency on centralized media distributors such as news platforms or social media sites such as YouTube, Facebook, or X (formerly Twitter) to adopt content provenance standards is one of the current bottlenecks of wider adoption of C2PA. Illustrations of decentralized provenance tools are emerging as demonstrated by the work of Cross Platform Origin of Content (XPOC) for origin of content verification \cite{microsoft2023xpoc}. 

It remains to be explored how to incorporate provenance features into existing, widely-used generative AI models. Ideally, it would be possible to integrate these features into models without retraining them. In addition, it is unclear whether model developers have adequate incentives to incorporate such features into their models, and even if some regulated centralized models can be required to include provenance measures, other non-centralized models will still not incorporate them. Content provenance is not a silver bullet solution -- even systems that employ content provenance are susceptible to social engineering attacks. However, if a bad-actor were to attempt to re-purpose the metadata, the omission of a digital signature of a trusted public key weakens the success of this attack. One potential incentive for integrating provenance features might come from an increased user demand for verification of AI systems. This can be viewed similarly to the incentives to acquire HTTPS certificates for websites so users have confidence in interacting with websites. 

Throughout the history of the internet, content has existed largely detached from its original context, and content provenance tools can, accordingly, contain challenges to contextual confidence. The detachment of content from its source and modifiers undermines a participant's confidence in identifying the authentic context within which they are communicating. 

\begin{tcolorbox}[colback=gray!5!white,colframe=black!75!black]\footnotesize
\textbf{Challenges addressed:} impersonate the identity of an individual without their consent; create mimetic models; imitate members of specific social or cultural groups. 
\end{tcolorbox}

\subsubsection{Community Notes}

Community Notes is an open-source collaborative platform, currently implemented at X (formerly Twitter) that allows communities to add crowd-sourced context to specific pieces of digital information. It has been an active feature on X worldwide since December 2022, and its success has leaders in technology asking whether algorithms of a similar sort can be adopted in other settings \cite{Buterin2023}. The potential extension of Community Notes beyond X, particularly for crowd-sourced fact-checking, presents notable advantages for larger organizations. However, while promising, Community Notes is not a flawless solution for adding context to bits of communication. Its effectiveness is tempered by limitations in response speed, underscoring the need for complementary strategies in information validation processes \cite{Bloomberg}. 

Community Notes is designed to counteract the prevailing, context-free norms of the internet, where information is often presented out of context. It does this by enabling a subset of X users to actively participate in fact-checking and context provision. This reinforces the idea that every piece of content should be subjected to questioning and verification through broad consensus. 

\begin{tcolorbox}[colback=gray!5!white,colframe=black!75!black]\footnotesize
\textbf{Challenges addressed:} falsely represent grassroots support; impersonate the identity of an individual; imitate members of social or cultural groups; disseminate content in a scalable, automated, and targeted way; fail to accurately represent the origin of content; misrepresent members of specific social or cultural groups.    
\end{tcolorbox}

\subsubsection{Centralized digital identities} \label{sec:centralized_id}

Centralized digital identities are digital representations or attestations of a participant's identity, issued by a centralized authority. Two common issuers of centralized digital identities are government actors and private companies. 

Digital government attestations can include digital birth certificates, health cards, driver’s licenses, passports, and voter cards. Many countries currently use government-issued digital identities to access various government services (Estonia \cite{eestonia2023eidentity}, India \cite{uidai2023getaadhaar} and Singapore \cite{singpass2023digitalid} are prominent examples). These digital identity solutions rely on cryptographic primitives (e.g. selective disclosure JWT \cite{ietf-oauth-selective-disclosure-jwt-2023}, U-Prove \cite{microsoft2023uproject}, BBS+ signatures  \cite{bbs-signature-scheme-2023}) that ensure one attestation cannot be associated or traced back to another such as verifiable credentials or mobile driver's licences \cite{w3c2022vcdata,aamva2023mdl, dgx2022digitalidentity}. 

Private companies have also become a primary issuer of digital identities. For example, it is common to sign into various third party sites with Google, Microsoft, or Facebook credentials as a form of authentication. The number of non-government issued digital identity solutions is growing – teams at Apple \cite{apple2023visionpro}, LinkedIn \cite{microsoft2023linkedinentra}, and Worldcoin \cite{worldcoin2023worldid}, for example, have each launched their own digital identity solutions in the last year. 

When participants use centralized digital identities to establish their identity online, their communication partners can verifiably identify displayed attributes of the person with whom they are communicating. Thus, centrally issued-digital identities help to identify context – they could be used to “tag” outgoing communications from a particular person or collective, whether on email, phone, text, social media or the internet. Centralized digital identities aim to contain a challenge to contextual confidence in that they override the norms of anonymity or pseudonymity that typically reign on the internet. 

The perspective of contextual confidence also helps to illuminate the downsides of centralized digital identity solutions. The centralization of power, either in a government or a corporation, raises concerns about the extent to which information be used by the centralized actor outside the intended context. Consider, for instance, India’s national digital identity system, Aadhaar. Signing up for Aadhaar was presented as optional, but it was also made to be a prerequisite for filing taxes. Given the way in which Aadhaar tracks individuals’ activities across a wide range of contexts, many have voiced concerns about surveillance and the vulnerability of this database to attacks \cite{wired2023aadhaar}. 

Or, in the non-government case, consider, Worldcoin – a project whose mission is to be a global issuer of digital identity. Worldcoin's governance structure reflects significant centralization: a small group controls a large portion of the decision-making power \cite{worldcoin2023whitepaper}. When checks on the power of a centralized issuer of identity are limited, and when the issuer is operating supranationally, the work that standards bodies have done to enhance individual control of information (e.g. GDPR and restrictions on third-party cookies) could be undermined.

\begin{tcolorbox}[colback=gray!5!white,colframe=black!75!black]\footnotesize
\textbf{Challenges addressed:} impersonate the identity of an individual without their consent; create mimetic models; imitate members of specific social or cultural groups; falsely represent grassroots support or consensus (astroturfing). 
\end{tcolorbox}

\subsubsection{Identity as a social intersection} \label{sec:social_id}

To avoid the centralization of power in a single issuer of digital identity, many have proposed digital identity solutions that harness social networks as a source of verification. Verifying identity as a social intersection is the process of identifying and authenticating participants through a set of attestations. The set of attestations do not have to be  rooted in centralized authorities, but can derive from any context an actor belongs to, including academic institutions, workplaces, social circles or friendships, for example. This idea is anchored in the premise that human identities are inherently social \cite{immorlica2019verifying}.

The most common example of verifying through a social intersection is the Open Authorization (OAuth) protocol \cite{ietf2023oauth}. OAuth is the most popular open authentication system for federated identity, and allows users to choose from a wide range of providers in order to access third-party applications. By “signing-in” with their Google or X (formerly Twitter) account, users can bootstrap their identity without providing any new information about themselves.  Gitcoin Passport is another an early example of an identity aggregator \cite{gitcoin2023passport}, along with other decentralized identity projects like Ethereum Name Service \cite{ens2023domains} or Spruce ID \cite{spruceid2023}, or “proof of personhood” protocols like Proof of Humanity \cite{proofofhumanity2023} or WorldID \cite{worldcoin2023worldid}.  Many identity protocols have their own authentication strategies. These strategies are surveyed in \cite{siddarth2020watches} and \cite{jain2022plural}. 

By approaching identity verification as a social intersection, we can adapt both the authentication process and the methods used to ``tag" communication outputs according to the specific context. This method contains challenges to contextual confidence, by superseding the prevailing culture of pseudonymity on the internet. 

There are many open questions about how to implement a social identity verification system, which the contextual confidence perspective throws into relief. For example, should all social attestations be equally valuable? In a communication landscape where AI is prevalent, it may become easier to subvert these social attestation systems through AI-generated attestations or AI-enabled collusion. We discuss collusion-resistant identity systems in section \ref{sec:collusion-identity}.

Finally, we acknowledge here that there are some benefits to preserving spheres of anonymous or pseudonymous communication -- anonymity can encourage openness and transparency about topics that would be sensitive to discuss under conditions of exact identification, and can, relatedly, protect speakers from retaliation or stigma. So, whether centralized or not, digital identity systems need not be required in all forms of communication.

\begin{tcolorbox}[colback=gray!5!white,colframe=black!75!black] \footnotesize
\textbf{Challenges addressed:} impersonate the identity of an individual without their consent; create mimetic models; imitate members of specific social or cultural groups; falsely represent grassroots support or consensus. 
\end{tcolorbox}

\subsection{Mobilization Strategies to Identify Context}

\subsubsection{Watermarking}

Watermarking is a technique intended to allow for disclosure and detection of the use of an AI model. It works by embedding a hidden pattern or ``watermark" into digital content, such as text, images, or videos, that is typically imperceptible to humans but can be algorithmically detected.

There are many different proposed approaches to watermarking for AI models.\footnote{For a few recent proposals, see \cite{wen2023tree, kirchenbauer2023watermark, abdelnabi2021adversarial, zhao2023provable, aaronson2023aisafety, gowal2023identifying, meta2023stablesignature}.} The goal of watermarking is to enable receivers to identify whether some content has been generated by a specific model. The hope is that such techniques will help to prevent various forms of deceptive misuse of AI models. There are many obstacles in the way of successful and reliable watermarking for AI models.\footnote{The main concerns around watermarking are its robustness and susceptibility to evasion \cite{jiang2023evading, zhao2023generative, kirchenbauer2023reliability}. In addition, watermarking can only be enforced in models that are centralized in their training and access. So even if watermarking technologies were to become robust and implemented in highly regulated and controlled models, the watermarking requirement could push malicious actors to substitute toward non-regulated and more harmful models.} Nonetheless, it is a promising technology for that may help to mobilize contextual confidence and set norms around disclosure and detection of AI model use \cite{shoker2023confidence}.

\begin{tcolorbox}[colback=gray!5!white,colframe=black!75!black]\footnotesize
\textbf{Challenges addressed:} impersonate the identity of an individual without their consent; create mimetic models; imitate another AI model. 
\end{tcolorbox}

\subsubsection{Model verification}
\label{sec:model_verification}

As generative AI models continue to proliferate – each uniquely characterized in terms of base weights – it will become important for users of these models to verify they are using the intended model, and not an imitation. When we discuss ``verifiable models," we are primarily focused on two areas of verification: (i) model weights, and (ii) model inference. We also discuss data verification in section \ref{sec:data_verification}. 

For proprietary models, model weights are protected within the context of the AI lab that developed the model. Although these weights have information security protocols in place to prevent them from being released \cite{karpur2023securing}, we often want to know that the same weights are being used across time and that we are dealing with the intended model and not an imitation. This fact can be proven to users by adding weight-based artifacts to an execution of a model (i.e., a model provider can include the hash of the model weights with each execution). Weight-based artifacts can help AI model developers to protect users from imitation models while still protecting the model’s sensitive information. This approach works well when model providers are trusted to faithfully report model weight derivatives (i.e., model signifiers). However, in contexts where vendors want true verifiability of model usage, we can draw on techniques that target  inference. 

Proving model inference relies on a set of zero-knowledge machine-learning (ZK-ML) tools \cite{kang2023verifiedexecution, kang2022scaling, ezkl2023} that enable the verification of provable claims such as “I ran this publicly available neural network on some private data and it produced this output” or  “I ran my private neural network on some public data and it produced this output” or “I correctly ran this publicly available network on some public data and it produced this output”.\footnote{These techniques potentially serve as valuable tools for evaluations, model cards, and audits, enabling labs to provide verifiable proof of these actions before releasing a model \cite{evals2023update, openai2023gpt4systemcard, anthropic2023claude2modelcard, cohere2023generation, mokander2023auditing, cihon2021ai, brundage2020toward, raji2020closing, katyal2019private, gebru2021datasheets, raji2022outsider}.} These claims are cryptographically secure proofs of inferences that can be cheaply verified by anyone without accessing original model weights. Currently, proving model inference using ZK-ML tools is possible, but still slow for large models. Researchers are actively working to improve the performance of ZK-ML tools for large models \cite{kang2023tensorplonk, zkml2023ezkl}. 

Model verification aims to mobilize contextual confidence by proactively addressing the absence of established norms for human-AI interactions. By enabling verifiable identification of intended AI models and inferences, model verification promotes contextual confidence in these novel interactions.

\begin{tcolorbox}[colback=gray!5!white,colframe=black!75!black]\footnotesize
\textbf{Challenges addressed:} imitate another AI model. 
\end{tcolorbox}

\subsubsection{Relational passwords}

As generative AI-enabled social engineering attacks proliferate, it will be important to have tools to bilaterally authenticate social relationships. Sophisticated scams that target close relationships -- i.e. when a parent receives an urgent call from a child in trouble, where the child's voice is cloned by an AI with high fidelity -- demand new forms of social authentication. What we call ``relational passwords" are passwords -- words or phrases or call-and-answer routines -- formed between two people who frequently communicate. When there is any suspicion that one might be speaking to an impersonator rather than the purported individual, the suspicious party can ask for the relational password. For instance, the parent could ask their child the name of their third grade teacher. These passwords can be used to thwart a wide range of social engineering attacks. 

The paradox of good passwords is that they need to be hard to guess but easy to remember. Relational passwords have the benefit of emerging out of a specific context (by being relational). A good relational password further takes advantage of context, drawing on the relationship itself to come up with something that is hard to guess yet easy to remember, such as a shared memory like ``The place we first met in the summer of 2002." For more intimate relationships, it is easier to come up with good relational passwords – they draw on shared history and experience. For more distant relationships (i.e. within and between organizations), it can be harder to come up with something that members know but a hacker cannot learn. 

Despite their many benefits, there is a tradeoff in making a relational password explicit -- a tradeoff between hackability and recall. The tradeoff between hackability and recall involves balancing the security risks of having a fixed, written-down password against the ease of remembering and using it. A predefined password, while easily recalled and convenient, is vulnerable to hacking, especially if stored digitally. On the other hand, relying on spontaneous, personal knowledge for authentication reduces the risk of being hacked, but can be challenging under stress or pressure, as it requires quick and clear recall of specific, potentially obscure details.

Digital authentication manifests in different forms as either something you know, something you are, or something you have. The norms regarding ``something you know" have traditionally been demonstrated through often ``contextless" compositions of characters and numbers or the frequent use of a single contextual ``recovery" password (e.g. mother's maiden name, elementary school, childhood street). Relational passwords aim to set higher expectations for the norms of ``something you know" by embedding context into what each of these ``passwords" unlocks, mobilizing contextual confidence in a communication landscape where generative AI is prevalent.

\begin{tcolorbox}[colback=gray!5!white,colframe=black!75!black]\footnotesize
\textbf{Challenges addressed:} impersonate the identity of an individual without their consent; create mimetic models; imitate members of specific social or cultural groups.  
\end{tcolorbox}

\subsubsection{Collusion-resistant digital identities}
\label{sec:collusion-identity}

As mentioned in section \ref{sec:social_id}, social verification of identity raises many opportunities for malicious influence, especially in a communication landscape where AI is prevalent. Generative AI can help dishonest actors subvert the system either by creating false attestations, or by convincing authentic participants to give inauthentic accounts false attestations. Unlike other identity systems, it is not just the volume of attestations that matter but rather the breadth of their social origin. Multiple attestations from a singular social circle might indicate potential collusion, whereas diverse sources of attestations often contribute to a more robust authentication. Collusion-resistant digital identity schemes aim to calibrate an actor’s influence within a given social context. Collusion-resistant identity solutions thus address the biases that can arise when all attestations are treated equally. This approach discounts influence in situations where there is a high degree of correlation within a context to mitigate collusion \cite{weyl2022decentralized}. Early implementations of collusion-resistant identities are illustrated in the context of voting \cite{ethresearch2023pairwise} and communication channels \cite{pluralcc2023github}. 

There are many open questions about how to discount influence. While a collusion-resistant digital identity system's design aims for robust authentication through diverse attestations, if the precise mechanisms for discounting influence are disclosed, technically sophisticated actors may be able to circumvent the protocols.\footnote{The tradeoffs associated with disclosing collusion-resistant digital identity schemes are discussed in \cite{weyl2022decentralized}.} Collusion-resistant identities aim to mobilize contextual confidence by establishing new strategies that incorporate the quantity of social intersections as part of the verification process, anticipating new challenges that generative AI mounts against social identity solutions. 

\begin{tcolorbox}[colback=gray!5!white,colframe=black!75!black]\footnotesize
\textbf{Challenges addressed:} falsely represent grassroots support or consensus (astroturfing); imitate members of specific social or cultural groups; impersonate the identity of an individual without their consent.  
\end{tcolorbox}

\subsection{Containment Strategies to Protect Context}

\subsubsection{Usage and content policies}

Usage and content policies can range from defining which users can access a specific model (who, where, when) or system, to describing and enforcing ``appropriate'' or ``allowed" uses of that model or system (how). Specialized models, such as those developed for sectors like the military or healthcare, might have unique access prerequisites. These prerequisites could include specific logins, clearances, or tailored classifiers that guide the model's application. Additionally, access might be influenced by users' expectations regarding how their interaction with the model will be monitored.\footnote{For a few recent approaches to AI licensing and structured access of AI models, see \cite{shevlane2022structured, anderljung2023protecting, anderljung2023frontier} and \cite{solaiman2023gradient}.} In AI deployment, model access is sometimes restricted based on a user’s IP address \cite{anthropic2023claude2} or subscription licenses \cite{openai2023gpt4}. Other model deployments require professional credentials, such as clinician status to access a medical model \cite{singhal2022large, googlecloud2023medpalm}.

Despite their benefits, usage policies also have several downsides. Usage policies can be abused. In some cases, they may lead to inequitable differences in access that must be weighed against the benefits of safe, restricted access.\footnote{For a critical discussion of usage policies in response to \cite{anderljung2023frontier}, see \cite{fastai2023dislightenment}.} In addition, usage policies are most easily employed in models that are centrally controlled---for open access models, it is harder if not impossible to enforce usage policies. 

Usage policies may contain challenges to contextual confidence by bringing offline tools -- identity checks, knowledge restrictions, content ratings, to AI systems. Historically, a significant portion of the internet was openly accessible, allowing universal access to tools, platforms, and websites. However, such unrestricted exploration occasionally resulted in potential misuse.

\begin{tcolorbox}[colback=gray!5!white,colframe=black!75!black]\footnotesize
\textbf{Challenges addressed:} grant access to a context-specific model without context-specific restrictions; misrepresent members of specific social or cultural groups.  
\end{tcolorbox}

\subsubsection{Rate-limiting communication}

Rate-limiting policies, typically deployed by a platform or other digital service provider, impose a cost on information flowing to or from a user. These costs may be measured in money or time, depending on what is more feasible or effective in a given context. Rate-limiting defends against high-volumes of content being shared from a single source. Examples include transaction fees on blockchain networks \cite{nakamoto2008bitcoin}, viewing limits on social media platforms \cite{musk2023tweet}, and spam prevention protocols in email  \cite{googlesupport2023preventspam}. 

Rate limiting is not applicable to all AI models, and can be abused. Rate limiting, like usage policies and prompt interface design, will be largely employed in models that are centrally controlled. When improperly calibrated by the centralized authority, rate limits can hinder genuine communication or favor actors with abundant resources, inadvertently deterring participation and obstructing the open exchange of information traditionally associated with the internet. When properly tuned to the setting, rate-limiting can contain challenges to contextual confidence initiated by the early internet culture’s dogmatic pursuit of the ``freedom of information." 

\begin{tcolorbox}[colback=gray!5!white,colframe=black!75!black]\footnotesize
\textbf{Challenges addressed:} disseminate content in a scalable, automated, and targeted way; pollute the data commons. 
\end{tcolorbox}

\subsubsection{Prompt protection and interface design}

Many applications of generative AI will continue to feature prompts from a user fed directly into a model. There are at least three types of design choices around the prompting process that can be used to protect contextual confidence. 

The first set of design choices concern data loss prevention (DLP) techniques. DLP techniques can be enforced within prompts to protect information from being misused outside its intended context. For example, it would be desirable and fairly straightforward to impose policies on general purpose models that restrict sharing of Social Security Numbers or API keys \cite{microsoft2023dlp}. 

Second, the organization deploying the model can carefully design interface instructions to remind users of context within a given prompt \cite{openai2023custominstructions}. Prompt examples and guided templates can be helpful in this regard. In addition, notifications that periodically remind users of key features of the Terms of Services are important – i.e. users should be frequently reminded about how the information they provide to the model will or will not be used or shared by the organization deploying the model or any third party. 

Lastly, clear interface design is key to helping users understand a generative AI models capabilities \cite{petridis2023constitutionmaker}. Incorporating custom instructions or prompt templates are just two ways in which users can be guided to clarify context, enabling the model to respond more effectively. Additionally, when models proactively seek context or clarification, it not only refines the response but also reminds users they're interacting with a tool, not a human.

Despite the many benefits of prompt protection and interface design, some critics see these approaches as paternalistic. In addition, this design dimension is only relevant for models that are centrally controlled and accessed. 

These strategies for prompt protection and interface design aim to contain challenges to contextual confidence by superseding the internet norm of unrestricted prompt, message or search interfaces, and sometimes obscured Terms of Service.

\begin{tcolorbox}[colback=gray!5!white,colframe=black!75!black]\footnotesize
\textbf{Challenges addressed:} misrepresent members of specific social or cultural groups; leak or infer information about specific individuals or groups without their consent.  
\end{tcolorbox}

\subsubsection{Deniable and disappearing messages}

Deniable messaging is a cryptographic technique that allows for messages to be shared in a way that, if taken out of context, their authenticity cannot be definitively proven. Disappearing messages are messages that automatically erase after a set period of time, limiting the period of time over which a particular claim can be verified. 

Deniable messages have two cryptographic properties \cite{jakobsson1996designated}: (i) \textit{unforgeability} ensures the receiver is confident that the message genuinely originated from the designated sender, and (ii) \textit{deniability} guarantees that even if the receiver is certain about the origin of a message, they cannot later prove to others that it was sent by the purported sender. The power of deniable messages depends crucially on defining a set of ``designated verifiers" for each message. This unique attribute of deniable messages means that, while the communication is credible and trustworthy between the sender and the designated verifier, it lacks the conventional cryptographic assurance that would enable the verifier to demonstrate its authenticity to third parties, thereby upholding the essence of deniability in sensitive communications.

As persuasive machine-generated content of questionable authenticity proliferates, individuals will need to rely more on verification to provide credibility in order to take action on information. In addition, deniable messages are valuable within the data collection and model outputs phases of AI development. As more of the open internet continues to become flooded with AI generated content, it is critical for AI labs to protect their models from model collapse and train only on certain types of content of verifiable origin \cite{shumailov2023curse}. Deniable messages can also be a tool for the enforcement of AI developers' usage policies – the usage policy can dictate not only access but also restrictions on who can verify model outputs in particular contexts. A limitation of deniable messages is that it assumes a zero-trust relationship between the receiver and the third party. 

Additionally, addressing social media's dilemmas of trust, content overflow, and harmful influence, some have suggested that user-formed groups can be a solution \cite{lanier}. Incorporating designated verifier signatures would enforce that only individuals within these groups can authenticate messages, preventing outsiders from validating content out of context, further protecting context in the interaction. Deniable messages thereby contain challenges to contextual confidence by overriding the internet norm of information having inconsistent verifiability standards.

\begin{tcolorbox}[colback=gray!5!white,colframe=black!75!black]
\footnotesize \textbf{Challenges addressed:} leak or infer information about specific individuals or groups without their consent; pollute the data commons; grant access to a context-specific model without context-specific restrictions.   
\end{tcolorbox}

\subsection{Mobilization Strategies to Protect Context}

\subsubsection{Contextual training}

Generative AI models typically undergo two main training phases: pre-training and fine-tuning. During the pre-training phase, the model is exposed to a diverse set of data, but this data lacks task-specific, subject-specific, or context-specific nuances. The fine-tuning phase refines this base model, introducing more specific details and nuances, often tailored to a particular application or context. One prominent technique in this phase is Reinforcement Learning with Human Feedback (RLHF) \cite{ouyang2022training, bai2022constitutional, khani2023collaborative, openai2023democratic}.

Contextual training is an umbrella term we use to refer to context-specific (as opposed to task-specific) pre- and post-training. Consider a query such as ``Are these symptoms COVID?" A task-specific model might be hard-coded to provide a set answer, regardless of the user's intent. However, the ideal response could diverge dramatically based on the user's context. For a scriptwriter crafting a screenplay, they might be seeking dramatic or fictional symptoms for plot purposes. In contrast, an individual inquiring for medical advice may require evidence-based information. With task-specific training, the AI is essentially limited to a narrow distribution of predefined responses, whereas context-specific training – facilitated by custom instructions, system messages, or even real-world user feedback – enables the model to gauge and respond to a broader array of nuances, leading to richer and more relevant outputs. 

Contextual training tools help to mitigate challenges to contextual confidence that result from the training process of an AI model – when the AI model does not appropriately account for context in its inputs, it will struggle to appropriately account for context in its outputs.\footnote{To understand the importance of context in training, consider the release of Llama2, which demonstrated the importance of the \textit{quality} rather than the \textit{quantity} of data annotations in the fine-tuning of transformer-based AI models \cite{touvron2023llama}.} In order to assess whether a model can be appropriately deployed in a given context, it is important to understand how it was trained. However, unlike pre-training, where there may be more transparency of data mixtures in model cards, annotations used for supervised fine-tuning are closely guarded by AI developers.\footnote{Fine-tuning base models via supervised learning has become a common trend leading to the development of a suite of very powerful models such as Phi \cite{li2023textbooks}, WizardLM \cite{xu2023wizardlm}, XGEN \cite{salesforce_xgen}, Vicuna \cite{vicuna2023}, Falcon \cite{falcon_llm}, and Alpaca \cite{crfm2023alpaca}.} Combining increased transparency around training with techniques like content filtering or ``unlearning" that limit model outputs on areas where the model has less context could vastly improve contextual confidence in AI-enabled communications \cite{microsoft2023harrypotter}. A limitation of this approach is the potential cost of training this model and protecting the fine-tuned version of the models weights from being released.

Norms on how we engage various stakeholders into the development process of AI technologies, and incorporate that data prior to deployment remain immature. Additionally, there are many open questions around how we align AI systems to the preferences expressed by a specific context. Contextual training aims to mobilize contextual confidence by establishing new higher expectations for protecting context through the model development process. 

\begin{tcolorbox}[colback=gray!5!white,colframe=black!75!black]
\footnotesize\textbf{Challenges addressed:} fail to accurately represent the origin of content; misrepresent members of specific social or cultural groups. 
\end{tcolorbox}

\subsubsection{Data verification}
\label{sec:data_verification}

Data verification is the process of verifying a data mixture claimed to have been used at a given stage of training, ensuring that the data used was legitimately sourced from the stated providers. Despite the progress being made in ZK-ML as discussed in section \ref{sec:model_verification}, research on verifiable training proofs (i.e. confirming that specific training data was used in training a model) is nascent \cite{choi2023tools}. Currently, model developers could in theory make verifiable attestations about the data mixtures  used throughout model development, showing that the data come from licensed sources. However, model developers typically do not use these verification tools – instead, in model cards, the developers tend to make general and unverifiable claims about the data used in training, especially in fine-tuning, reward modeling, and reinforcement learning.

Consider the implications: should a clinician trust a model if they are unsure about whether the model was trained with credible clinical content? If an educational board is uncertain about the authenticity of a model's training material, would they introduce it as an AI tutor? Data verification serves as a tool to instill confidence that an AI model possesses the necessary expertise or cultural comprehension for its intended deployment \cite{longpredata}. 

Data verification mobilizes contextual confidence by illustrating exactly how information that originated in a particular context is repurposed by an AI model in a new context. 

\begin{tcolorbox}[colback=gray!5!white,colframe=black!75!black]
\footnotesize\textbf{Challenges addressed:} fail to accurately represent the origin of content; misrepresent members of specific social or cultural groups. 
\end{tcolorbox}

\subsubsection{Data cooperatives}\label{sec:datacoops}

Data cooperatives are member-governed entities that aggregate subgroups data and negotiate its usage guidelines, ensuring that the benefits derived from this shared resource are returned to the subgroups.\footnote{For agenda-setting treatments of data cooperatives, see \cite{hardjono2019data}, \cite{schwab2011personal} and \cite{radicalxchange2023datafreedomact}.} The biggest obstacle to such approaches is figuring out how to attribute value to subgroups. Recently developed methods such as influence functions \cite{koh2017understanding,feldman2020neural, grosse2023studying}, TRAK \cite{park2023trak}, datamodels \cite{ilyas2022datamodels} and Shapley values \cite{ghorbani2019data,jia2019towards} are promising, and could serve as the basis through which data cooperatives pass value back to their members.\footnote{For a survey of data attribution tools, see \cite{hammoudeh2022training}.}  

Data cooperatives can mobilize contextual confidence, seeing in the rise of generative AI an opportunity to set higher standards for data attribution, responsibility and monetization.

\begin{tcolorbox}[colback=gray!5!white,colframe=black!75!black]
\footnotesize\textbf{Challenges addressed:} fail to accurately represent the origin of content; misrepresent members of specific social or cultural groups.  
\end{tcolorbox}

\subsubsection{Secure data sharing mechanisms}

Many of the strategies for protecting context discussed above by definition inhibit data sharing across contexts. Nonetheless, there are many settings in which it is important to share data across contexts, revealing only those pieces of information necessary to the task at hand. Secure data sharing mechanisms will be an important tool for making sure that relevant information can still be shared while protecting context through the other strategies discussed above.

Secure data sharing mechanisms include cryptographic techniques like traditional encryption schemes, secure multiparty computation \cite{bogdanov2014input}, differential privacy \cite{dwork2006differential, abadi2016deep}, homomorphic encryption \cite{gentry2009fully}, secret sharing \cite{shamir1979share}, trusted execution environments \cite{sabt2015trusted}, as well as decentralized and federated training of models \cite{mcmahan2017communication}.

Consider a network of hospitals. Even if each hospital has the tools to protect its patient data with various other strategies discussed in this paper, there is a remaining challenge: How does each hospital share data with another hospital or AI model developer without undoing the protections it has applied? Secure data sharing mechanisms can facilitate the sharing and networking of contexts while ensuring high contextual confidence. The theoretical ideas behind the secure data sharing mechanisms we discuss here have been around for a long time, but they are relatively untested in the practice. In order to make these mechanisms practical, more empirical research is needed.

While regulations such as HIPAA address data protection within the healthcare domain, there remains significant ambiguity around broader norms for secure data sharing across contexts in the presence of generative AI. By proactively implementing secure data sharing mechanisms, we can mobilize contextual confidence across interconnected contexts. 

\begin{tcolorbox}[colback=gray!5!white,colframe=black!75!black]
\footnotesize\textbf{Challenges addressed:} Leak or infer information about specific individuals or groups without their consent. 
\end{tcolorbox}

\subsection{Implementation of Strategies to Promote Contextual Confidence}

We have now discussed a number of strategies that promote contextual confidence in communications. In this subsection, we clarify how these strategies relate to each other, and highlight the roles of different actors in their development. 

What we give now is a crude outline of the chain of implementation of the strategies considered above, focusing on what delivers contextual confidence in communication technologies. We stress again that there are many important strategies for promoting contextual confidence not discussed here, and that there are considerations (especially around hardware, for example) that lay outside the scope of the present discussion. We also discuss only a subset of actors, mainly those who can develop and deploy these technologies in the near term, leaving aside the question of how these strategies can be strengthened through broad coordination e.g. government regulation, international ethics committees, and industry-wide standards organizations.\footnote{For further discussion on the roles different actors play in responsible AI deployment see \cite{SafetyWhiteHouse, ho2023international, schuett2023towards, whitehouse2023ai} and \cite{anthropic2023responsiblescaling}.}

Communication technologies, which form the foundational layer, are systems used to send and receive information. These include email and messaging services, social media platforms, news platforms, and other communication platforms for cultural production and consumption. In order to ensure contextual confidence in this root layer, there must be identity protocols and model development policies in place beforehand. Through identity protocols, users gain access to communication technologies, and the details of the identity protocol determine norms of communication.\footnote{For example, if a web application requires only a valid email address and allows for pseudonymous user names, the norms of communication will be different from an organization-wide Slack accessible only to verified members of the organization.} As generative AI becomes integral to communication, choices made at the AI model development layer also crucially feed into the communication layer. As messages flow between users, data management protocols that sit on top of the communication protocol can ensure context is protected and appropriately shared back to AI model developers. 

So, there are four key layers of implementation: AI model development, identity protocols, messaging protocols, and data management protocols. It is at the layer of AI model development where usage policies, model verification, watermarking, data verification and contextual training are most likely to be effectively implemented. Meanwhile, the identity protocol layer is where digital identity solutions and relational passwords should be pursued. The communication technologies themselves can pursue strategies such as rate limits, content provenance, Community Notes, deniable messages and prompt design. Finally, data management protocols such as data cooperatives and secure data-sharing mechanisms protect contextual confidence on the communication layer, and protect context as information feeds back into AI model development.

\begin{figure*}
    \centering
    \includegraphics[scale=0.4]{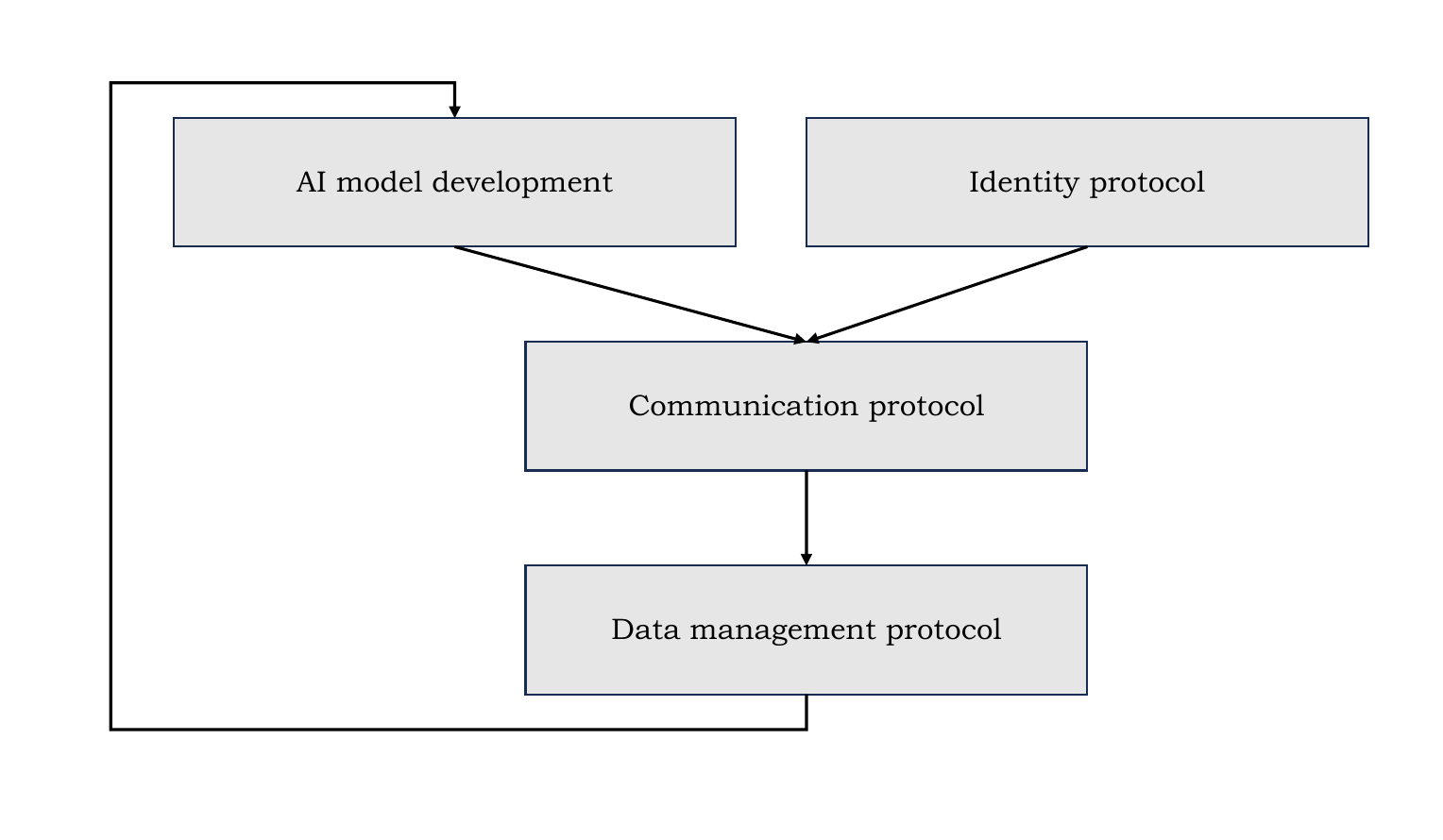}
    \caption{Summary of parties to pursue strategies discussed in this report.}
    \label{fig:actors}
\end{figure*}

\subsection{Illustrating the Value of the Contextual Confidence Perspective} 

In this section, we show how the contextual confidence framework helps to guide action in a specific, concrete setting. We contrast the approach suggested by contextual confidence with the approaches suggested by perspectives focused on  \textit{privacy} and \textit{information integrity}. 

Suppose a company’s CEO is interested in developing a mimetic model of herself. The purpose of this model is to maximize the CEO's ability to field concerns raised by employees, to serve as a ``proxy CEO" by attending meetings on the CEO's behalf, and to gather information in order to provide advisory support to the human CEO. In this discussion, we will compare the recommendations that might emerge from a conventional privacy or information integrity review of such a tool with a review focused on securing contextual confidence. We hope this case illustrates that, compared to privacy and information integrity perspectives, contextual confidence suggests an approach that mitigates risks of deploying this proxy CEO model, without sacrificing too much of its usefulness. 

We first consider a ``privacy" perspective, and in particular a perspective that sees privacy as secrecy about and control over one's personal data.\footnote{Of course there are many possible conceptions of privacy not captured by this one, but we take this one as an influential example.} From this type of privacy perspective, the recommendations would likely focus on the removal of sensitive data from both the training data and outputs of the mimetic model. This could involve employing data loss prevention or de-identification techniques to either transform or completely remove sensitive information. This privacy-as-control approach certainly gets some things right. However, at the same time, it is hard to know where to draw the line on what constitutes sensitive information, and there is a risk that an overly conservative approach might strip away the very data that makes the mimetic model effective. There is a deep tension between privacy and functionality here, and the perspective of privacy as secrecy of personal data does not offer much useful guidance as to how to strike the right balance.

On the other hand, an information integrity perspective would advocate for a pursuit of ``truth," striving to, in some sense, ``fact-check" model outputs. That is, this approach might lead to the development of a high frequency review process for the mimetic model's outputs, with a goal of ensuring that the model's responses are in some sense an accurate proxy for the CEO's putative responses. Again, like the privacy-as-control approach, the information integrity approach also serves as a good start. However, in a large company, the review process may not scale and may be overburdening, especially for the CEO who created the model in the first place to improve efficiency in her communications. 

How does contextual confidence guide action in this scenario? Let us first evaluate the challenges to the identification and protection of context in this scenario. The CEO model introduces uncertainty into employees’ ability to accurately identify whether they are interacting with the authentic CEO or the authentic CEO model (could be a malicious model) they are interacting with. Additionally, the norms governing the interaction – and how the information exchanged might subsequently be utilized by the company, the CEO, or other employees remain ambiguous. To navigate these complexities, a framework grounded in contextual confidence would recommend at least four specific actions: 

\begin{enumerate}
    \item \textbf{Usage policies}: Require employees to pass comprehension tests to engage with CEO model. 
    \item \textbf{Model verification:} Develop proofs that can be widely used to regularly verify CEO model.
    \item \textbf{Interface design}: Disclose model use in all outbound communications. 
    \item \textbf{Deniable messages}: Deploy designated verified signatures for off-the-record conversations. 
\end{enumerate}

Through a hypothetical example of a company's CEO developing a mimetic model to enhance efficiency, this section aimed to concretely illustrate the value of the contextual confidence perspective. While more conventional \textit{privacy} (as control over personal information) and \textit{information integrity} perspectives offer valuable insights, they may not always strike the most effective balance between protection and functionality. The contextual confidence framework, however, offers a nuanced approach that emphasizes the importance of recognizing and safeguarding context. The framework suggests a set of specific actionable recommendations, from setting usage policies and model verification strategies to refining interface designs and ensuring that there are deniable messaging protocols to enable genuinely off-the-record conversations. These recommendations underscore the potential of the contextual confidence approach in offering a balanced, effective, and comprehensive strategy for navigating emerging complexities around generative AI model use.

\section{Discussion}
\subsection{Enforcing Protective Norms}

In this paper, we have highlighted strategies that promote contextual confidence by setting new norms and expectations around the identification and protection of context. This discussion leaves out an important aspect of contextual confidence: It is important not only that there are norms, but that the norms are respected. Indeed, a norm cannot become a norm if there is no expectation that it is respected. 

There are two primary stages to any enforcement strategy to protect norms: \textit{commitment} strategies and \textit{accountability} strategies.  

\textit{Commitment} strategies aim to create shared knowledge of the context that requires protecting. For example, a participant in a video call might ``commit" to acknowledging that the meeting is being recorded, implicitly agreeing to the Terms of Service pertaining to that recording. Such a commitment engenders shared awareness among participants about two key aspects: the ongoing recording of the meeting and the terms governing the use of the information derived from it. Commitment in this example enhances the protection of contextual confidence from the default case in which  participants may inappropriately record conversations to then reuse or repurpose content from those conversations. However, commitment alone does not enforce the protection of context for participants who do not respect their commitments. 

 \textit{Accountability} strategies ensure that participants in a communicative exchange are held accountable if they violate their commitments. In some cases, the law  naturally provides accountability. In the Terms of Service example above, a violation of Terms of Service may be treated as a breach of contract, which can potentially be litigated in court. In other cases, the law will not as readily provide accountability, and specific organizations and platforms will need to develop their own mechanisms for accountability. 

Existing commitment and accountability strategies are sparse and require further development and experimentation. While some instant messaging platforms inform users if their communication partner has taken a screenshot of their chat, this simple commitment tool is far from standard or widespread. While some platforms have robust systems for reporting fake accounts, even these accountability technologies are often ineffective. While, as discussed above, prominent video calling services have deployed features that announce when the meeting is being recorded, individuals may not understand Terms of Service documents in full for each video call in which they participate. Thus, while there may be some commitment in the video calling case, this commitment may not be well understood.

\subsection{Open Questions and Future Work}

We hope this paper serves as a modest starting point for future collaborations between policymakers, AI model developers and researchers, focused on applying a contextual confidence perspective in particular domains. We outline here a few  areas of future work that may be especially valuable in the near-term.  

First, pragmatically defining what constitutes ``context" is challenging in some domains. The concept of context itself is somewhat slippery, and certain elements of context are more important than others in particular domains. Identifying the “who, why, where, when, and how” in every information flow may be impractical at some scales of applications, and this paper offered little guidance as to how to prioritize the most important elements of context. It would be useful to build toward a standardization of what qualifies as a comprehensive contextual confidence evaluation that could be incorporated into safety reviews, or even model cards. 

Second, we hope that this paper serves as a call to action for prioritizing the research and development of strategies that promote contextual confidence. Many of the strategies we discuss in this paper are in early stages of development and need a lot more research and development before they can be deployed. In addition, our enumeration of strategies is far from exhaustive. For example, we only offered cursory discussions of how AI models themselves can be used to strengthen strategies for promoting contextual confidence. 

Moreover, as noted throughout the discussion of the strategies in \autoref{sec:strategies}, many of the strategies we discuss here are effective only for models that are accessed through a single centralized entity. In particular, usage policies (III.C.1), rate limiting policies (III.C.2) and interface design (III.C.3) rely on centralized access, while contextual training (III.D.1) and to some extent watermarking (III.B.1) and content provenance (III.A.1) rely on centralized training. In a generative AI ecosystem dominated by open source models, many of the strategies we discuss will be moot or muted in their effectiveness. In future work, we hope to further explore strategies that are more directly targeted at increasing contextual confidence in the presence of open source models.

Third, it is critical to conduct empirical usability studies and surveys about whether and how the strategies we discussed indeed promote new norms in communication. Some of the strategies we discussed may have unforeseen consequence when applied in particular situations – like the surprising findings about how verification badges and fact checking tools may backfire in some contexts \cite{xiao2023account, akhawe2013alice}. In addition, it is important to gather data on the degree to which the strategies discussed here are differentially usable and accessible for different participants. 

In this paper, we focused on the ways generative AI challenges contextual confidence. As communication technologies continue to evolve, challenges to contextual confidence will continue to emerge beyond generative AI. For instance, advancements in augmented reality and robotics may bring a whole other set of difficulties into the identification and protection of context in the physical world. It is our hope that framing challenges to effective communication in terms of contextual confidence will be useful in forthcoming stages of technological development. 

\clearpage

\pagebreak 

\footnotesize
\section*{Acknowledgements}
We thank Glen Weyl, Danielle Allen, Allison Stanger, Miles Brundage, Sarah Kreps, Karen Easterbrook, Tobin South, Christian Paquin, Daniel Silver and Saffron Huang for comments and conversations that improved the paper.

\bibliographystyle{IEEEtran}  
\bibliography{references}  

\clearpage

\end{document}